\documentclass[conference,a4paper]{APSIPA2021}
\usepackage{amsmath}
\usepackage{graphicx}
\usepackage{multirow}
\usepackage{threeparttable}

\usepackage{bm}

\usepackage{geometry}
\usepackage{fancyhdr}

\fancypagestyle{firststyle}{
  \fancyhf{}
  \fancyhead[C]{2024 Asia Pacific Signal and Information Processing Association Annual Summit and Conference (APSIPA ASC)}
}

\begin{document}

\title{Multibiometrics Using a Single Face Image}

\author{
\authorblockN{
Koichi Ito\authorrefmark{1}, Taito Tonosaki\authorrefmark{1}, Takafumi Aoki\authorrefmark{1}, Tetsushi Ohki\authorrefmark{2}, and
Masakatsu Nishigaki\authorrefmark{2}
}

\authorblockA{
\authorrefmark{1}
Graduate School of Information Sciences, Tohoku University, Japan\\
E-mail: ito@aoki.ecei.tohoku.ac.jp}

\authorblockA{
\authorrefmark{2}
Faculty of Informatics, Shizuoka University, Japan}
}

\maketitle
\pagestyle{fancy}

\begin{abstract}
  Multibiometrics, which uses multiple biometric traits to improve recognition performance instead of using only one biometric trait to authenticate individuals, has been investigated.
  Previous studies have combined individually acquired biometric traits or have not fully considered the convenience of the system.
  Focusing on a single face image, we propose a novel multibiometric method that combines five biometric traits, i.e., face, iris, periocular, nose, eyebrow, that can be extracted from a single face image.
  The proposed method does not sacrifice the convenience of biometrics since only a single face image is used as input.
  Through a variety of experiments using the CASIA Iris Distance database, we demonstrate the effectiveness of the proposed multibiometrics method.
\end{abstract}
\section{Introduction}

Biometric recognition is a technology that identifies individuals using physical characteristics such as face, iris, and fingerprints, or behavioral characteristics such as gait and handwriting \cite{HoB}.
Biometric recognition is reliable and convenient compared to passwords and personal identification numbers, however, environmental changes such as brightness and pose may significantly degrade the recognition performance.
Therefore, multibiometrics, which uses multiple biometric traits to improve recognition performance instead of using only one biometric trait, has been investigated \cite{HoM}.

Multibiometrics improves recognition performance with more biometric traits to be combined.
For example, the combination of face, fingerprint, and iris recognition has been investigated \cite{Telgad-ICISIM-2017}.
Face recognition is highly convenient since the user only needs to turn his/her face to the camera to be authenticated, while the performance of face recognition is degraded by aging, cosmetics, and changes in the face pose.
Fingerprint and iris recognition require dedicated sensors, while their performance is more stable than that of face recognition.
Combining face, fingerprint, and iris recognition can improve the performance and stability of user authentication, however, the convenience is significantly reduced because of acquiring face, fingerprint, and iris at the same time.
In addition, the cost of the system increases since it ususally requires a camera for capturing the face, a sensor for capturing the fingerprint, and a dedicated camera for capturing the iris.
Other multibiometrics that combine iris and periocular recognition have been investigated \cite{Wang-TIFS-2021}.
Periocular recognition uses images of the eye and its surrounding area, and therefore the same camera can be used to capture both periocular and iris images.
Iris recognition has high performance, however, its performance depends on the accuracy of iris segmentation.
Periocular recognition has stable performance because of the use of periocular images, however, its performance is degraded by environmental changes.
Combining iris and periocular recognition improves the performance and stability due to their complementarity.
On the other hand, the distance between the user and the system may be limited due to the need to capture the periocular images.
Previous studies described above have combined individually acquired biometric traits or have not fully considered the convenience of the system.

\begin{figure}[t]
  \centering
  \includegraphics[width=.7\linewidth]{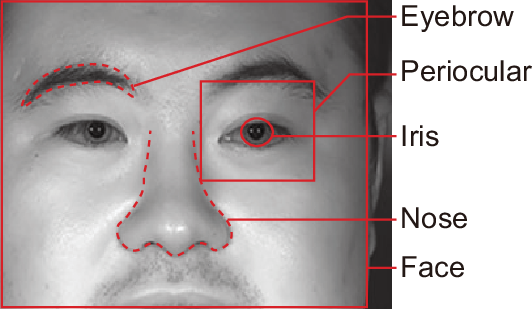}
  \caption{Biometric traits extracted from a single face image, whose combination is considered in this paper.}
  \label{fig:traits}
\end{figure}

Focusing on a single face image, we investigate a novel multibiometric method that combines five biometric traits, i.e., face, iris, periocular, nose, and eyebrow, that can be extracted from a single face image, as shown in Fig. \ref{fig:traits}.
The proposed method combines multiple biometric traits without sacrificing the convenience of biometrics since only a single face image is used as input.
To the best of our knowledge, there are no studies of multibiometrics, in which five biometric features are extracted from a single image.
In the proposed method, we first detect keypoints of a face using Mediapipe FaceMesh\footnote[1]{https://google.github.io/mediapipe} and obtain an image of each biometric trait based on these keypoints.
Next, we extract features using Convolutional Neural Network (CNN) and calculate the matching score based on features for each trait.
Then, we obtain the final matching score by fusing scores with a weighted sum.
Among the biometric databases available so far, only the CASIA Iris Distance database\footnote[2]{http://biometrics.idealtest.org} has a sufficiently large image size to extract images from face to iris.
Through a variety of experiments using the CASIA Iris Distance database, we demonstrate the effectiveness of the proposed multibiometrics method.

\section{Related Work}

The study of multibiometrics requires databases that consist of multiple biometric traits.
In the early days, XM2VTS \cite{Messer-AVBPA-1999} with face and voice and MCYT \cite{Ortega-IEE-2003} with fingerprint and signature have been usually used for the study.
NIST BSSR1 (Biometric Scores Set - Release 1)\footnote[3]{https://www.nist.gov/itl/iad/image-group/nist-biometric-scores-set-bssr1}, which consists of matching scores obtained from fingerprint and face recognition, has been also available.
In some cases, databases of a single biometric trait have been combined to create a pseudo multibiometric database for evaluation.
These databases are based on the assumption that biometric traits are acquired by multiple sensors, which sacrifices the convenience of biometrics even if it improves the recognition accuracy.
At that time, most studies have focused on investigating score fusion methods that improve recognition accuracy \cite{HoM}.

Recently, methods for multibiometrics using CNN have been proposed.
Stokkenes et al. proposed to combine face and periocular recognition at the score level \cite{Stokkenes-IPTA-2016}.
Telgad et al. proposed to combine three biometric features, i.e., face, fingerprint, and iris, at the score level \cite{Telgad-ICISIM-2017}.
Zhang et al. proposed to combine iris and periocular recognition using images captured by mobile devices \cite{Zhang-TIFS-2018}.
Wang et al. proposed to combine UniNet for iris recognition and AttNet for periocular recognition \cite{Wang-TIFS-2021}.
Ng et al. proposed CMB-Net, which combines face and periocular features \cite{Ng-ICPR-2022}.
In addition, there are other methods that combine the face, left iris and right iris \cite{Ammour-TSP-2018}, and the periocular and iris \cite{Wang-TIFS-2020,Tonosaki-APSIPA-2023}.
Since CNN can extract features with high discriminative performance and has improved the recognition accuracy of a single biometric trait, multibiometrics using CNN has also improved its recognition accuracy.
However, they only combine two or three biometric traits, or require multiple sensors.

\section{Multibiometrics Using a Single Face Image}

\begin{figure}[t]
  \centering
  \includegraphics[width=\linewidth]{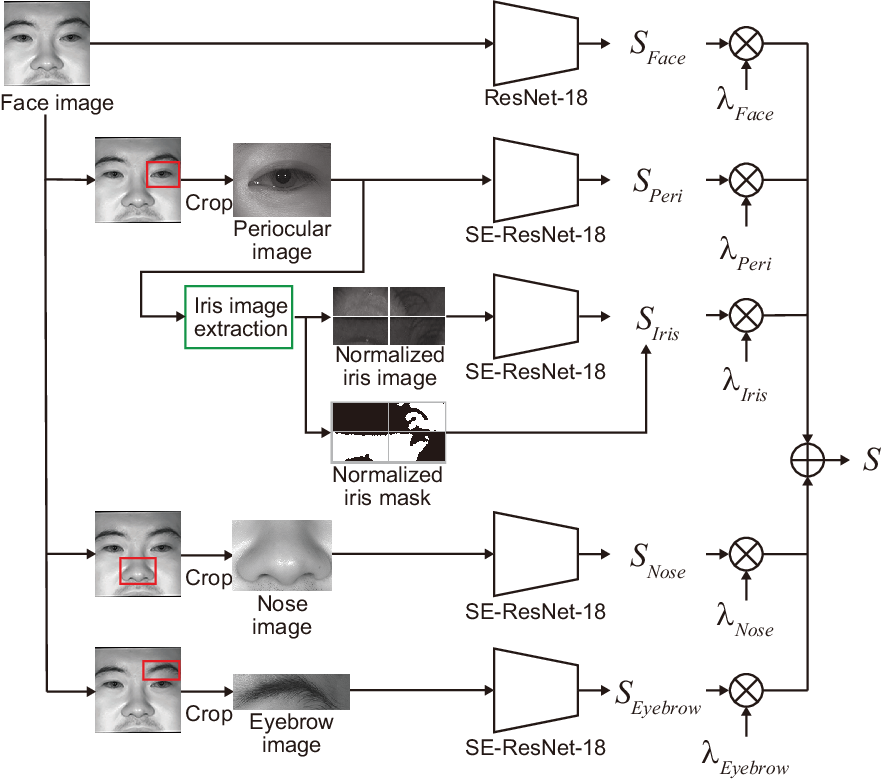}
  \caption{Overview of the proposed multibiometrics approach using a single face image.}
  \label{fig:proposed}
\end{figure}

We propose a multibiometric method that combines biometric traits extracted from a single face image: face, periocular, iris, nose, and eyebrow.
Fig. \ref{fig:proposed} shows an overview of the proposed method.
The proposed method consists of extraction of each biometric trait from a face image, feature extraction using CNN and calculation of a matching score for each trait, and score fusion by a weighted sum.
In the following, we describe the details of each process.

\subsection{Biometric Trait Extraction}

\begin{figure}[t]
  \centering
  \includegraphics[width=.5\linewidth]{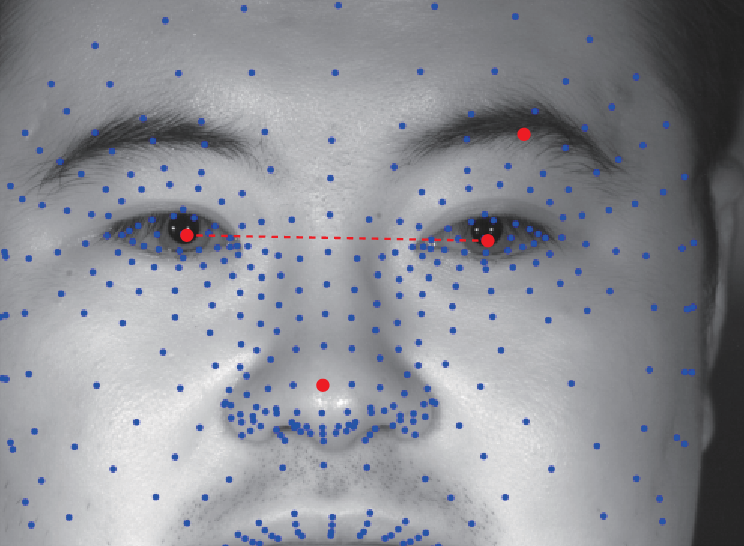}
  \caption{Keypoints extracted using Mediapipe FaceMesh.
  The proposed method uses the center of both eyes, the center of the nose, and the center of the left eyebrow among these key points, which are indicated by red-colored points.}
  \label{fig:keypoints}
\end{figure}

\begin{figure}[t]
  \centering
  \includegraphics[width=\linewidth]{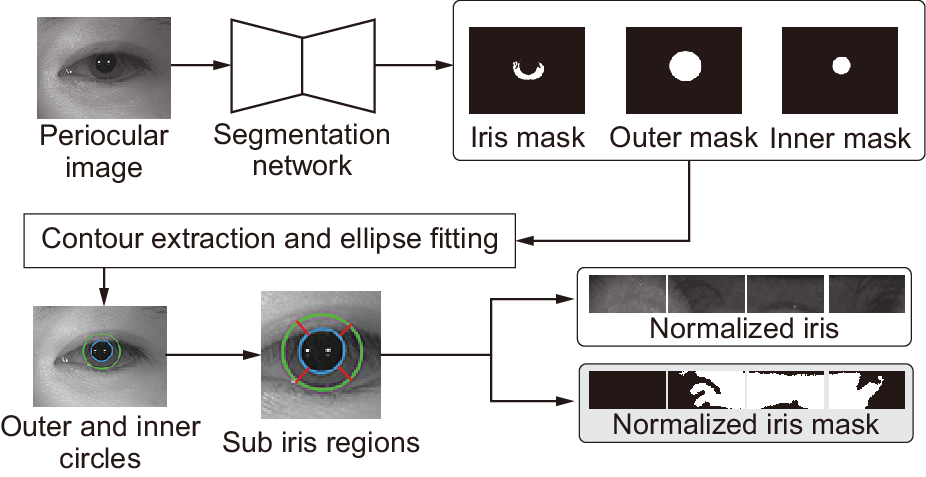}
  \caption{Flow of iris extraction from a periocular image \cite{Kawakami-APSIPA-2022}.}
  \label{fig:iris_extraction}
\end{figure}

This section describes the process of extracting biometric traits from face images.
We first detect keypoints on the face image using Mediapipe FaceMesh\footnotemark[1].
Fig. \ref{fig:keypoints} shows an example of the detected keypoints.
The proposed method uses the keypoints at the center of both eyes, the center of the nose, and the center of the left eyebrow among these keypoints.
Next, the rotation of the face image is normalized so that the line segment connecting the keypoints of both eyes is horizontal, and the normalized image is used as the face image for face recognition.
The center of the left eye is used to crop a rectangular region to obtain the periocular image used for periocular recognition.
Similarly, the center of the nose and the center of the left eyebrow are used to obtain the nose and eyebrow images for nose and eyebrow recognition, respectively.
The iris images are extracted from the periocular image using the method proposed by kawakami et al \cite{Kawakami-APSIPA-2022} as shown in Fig. \ref{fig:iris_extraction}.
The details of the process are omitted due to space limitation.
For more details, refer to \cite{Kawakami-APSIPA-2022}.
From the above process we obtain a face image, a periocular image, four normalized iris and mask images, a nose image, and an eyebrow image.


\subsection{Feature Extraction Using CNN}

\begin{figure}[t]
  \centering
  \includegraphics[width=\linewidth]{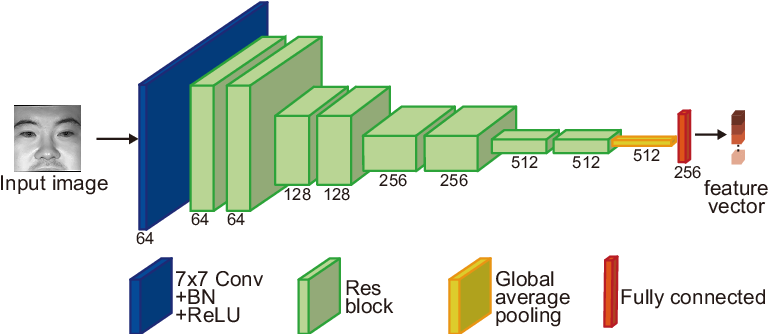}
  \caption{Network architecture of ResNet-18 for face recognition in the proposed method.}
  \label{fig:resnet}
\end{figure}

\begin{figure}[t]
  \centering
  \includegraphics[width=\linewidth]{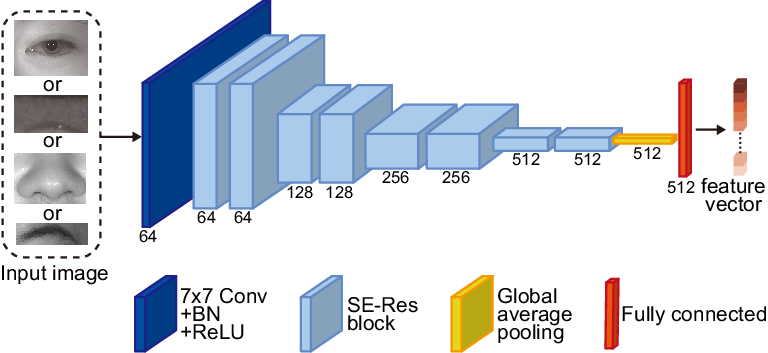}
  \caption{Network architecture of SE-ResNet-18 for periocular, iris, nose, and eyebrow recognition in the proposed method.}
  \label{fig:seresnet}
\end{figure}

We describe CNN for feature extraction from each biometric trait.
As for face recognition, we use ResNet-18 \cite{He-CVPR-2016} with ArcFace \cite{arcface} as shown in Fig. \ref{fig:resnet}, which is a popular approach in face recognition and achieves high recognition accuracy.
As for other biometric traits, we use SE-ResNet-18 \cite{He-CVPR-2016,Hu-CVPR-2018} as shown in Fig. \ref{fig:seresnet}.
SE-ResNet-18 is a CNN model that introduces a Squeeze-and-Excitation block (SE block), which is an attention mechanism, into ResNet.
In SE block, feature maps output from the convolution layer are averaged in the channel direction and aggregated into a vector, and then passed through two fully-connected layers and activation functions (ReLU and Sigmoid) to convert it into a vector that represents the weights for each channel.
The weight vectors are then multiplied by the feature map to obtain a refined feature map.
The Cross Entropy (CE) loss is used as the loss function for SE-ResNet-18 in the proposed method.
Note that for the iris image consisting of four subimages, we employ the method proposed by Kawakami et al. \cite{Kawakami-APSIPA-2022} for feature extraction.


\subsection{Score Fusion}

We describe a fusion approach of the matching scores obtained from each biometric trait.
In the proposed method, the matching score is calculated by the cosine similarity between the features extracted from each biometric trait using CNN.
As for iris recognition, we use a weighted average of the matching score of each subimage, where each weight is calculated based on their mask ratio, resulting in a score that gives more weight to subimages with less occlusion.

The final matching score is calculated by a weighed sum of all the matching score as follows
\begin{eqnarray}
  S &=& \lambda_{face}S_{face} + \lambda_{peri}S_{peri} + \lambda_{iris}S_{iris}\nonumber\\
  && + \lambda_{nose}S_{nose}+ \lambda_{eyebrow}S_{eyebrow},
\end{eqnarray}
where $S_\ast$ is the matching score for each biometric trait, and $\lambda_\ast$ is the weight of each, which satisfies the following equation
\begin{equation}
  \lambda_{face} + \lambda_{peri} + \lambda_{iris} + \lambda_{nose}+ \lambda_{eyebrow} = 1.
\end{equation}

\section{Experiments and Discussion}

In this section, we describe the evaluation experiments of the proposed method using the CASIA Iris Distance database\footnotemark[2].
In this experiment, we evaluate the recognition accuracy of each biometric trait alone and combinations of traits.
For the combination, we compare the proposed method with feature-level fusion in addition to the score-level fusion described above to demonstrate the effectiveness of the proposed method.

\subsection{Dataset}

In this experiment, we use the CASIA Iris Distance database\footnotemark[2].
This database consists of images taken of the upper part of faces of the subjects with a near-infrared camera from a distance of 2.4m$\sim$3.0m, and contains a total of 2,567 images taking from 142 faces.
Most of the images are blurred and do not show the iris clearly, making iris recognition particularly difficult.
In the experiments, we divide this database into three subsets: training, validation, and test, and use them to evaluate the recognition accuracy of single biometrics and multibiometrics.
The training data consists of 84 subjects (1,531 images) whose IDs are from S4000 to S4083, the validation data consists of 28 subjects (497 images) whose IDs are from S4084 to S4111, and the test data consists of 30 subjects (513 images) whose IDs are from S4112 to S4141.

\subsection{Evaluation Metrics}

We describe the evaluation metrics used in the experiments.
False Rejection Rate (FRR) is calculated from the matching score of genuine pairs and False Acceptance Rate (FAR) is calculated from the matching score of impostor pairs.
We use Equal Error Rate (EER), which is equal to FRR and FAR.
In practical authentication systems that require high security, such as immigration control systems, it is necessary to operate the system so that the FAR is low.
Therefore, we use FRR@FAR0.1\% and FRR@FAR0.01\%, which are FRR when the threshold is set so that FAR is 0.1\% and 0.01\%, respectively.
The smaller the value of EER and FRRs, the higher the recognition accuracy.

\subsection{Training Setting}

We describe the training settings of the network used in the proposed method.
ResNet-18 and SE-ResNet-18 are pre-trained using ImageNet and made fine-tuning using the training data of CASIA Iris Distance.
The initial value of the learning rate is set to 0.0001, and the learning rate is multiplied by 0.1 if the loss to the validation data does not improve for 10 consecutive epochs.
If the loss to the validation data does not improve for 30 consecutive epochs, training is terminated and the test data is evaluated using the weights from the epoch with the lowest loss.
The batch size is set to 4 and Adam \cite{Kingma-ICLR-2015} is used as optimizer.
When training CNN for periocular recognition, random crop, color jitter, random erasing, and random perspective are used for data augmentation and when training CNN for iris recognition, random erasing and random horizontal shift are used for data augmentation.
The random horizontal shift is a circular shift within a range of 10 pixels on the left and right sides, respectively.
All the data augmentation methods are applied with a probability of 50\%.
The segmentation network used for iris region extraction in Fig. \ref{fig:iris_extraction} is pre-trained using a subset of CASIA Iris Distance created by \cite{Wang-TIFS-2020}.
This subset is annotated with the ground truth labels of the iris, outer, and inner masks for iris segmentation.
The iris region is extracted by using the rectangle normalization module of OSIRIS \cite{Othman-PRL-2016}.

\begin{table}[!t]
  \caption{Results for a single biometric trait with different combination of CNN and its loss function, where R and S for CNN indicate ResNet-18 and SE-ResNet-18, respectively, and A and C for loss functions indicate ArcFace and CE, respectively.
  The underlines are the proposed combinations, and the bold numbers indicate the best for each biometric trait.
  The units for EER and FRR are percentages.}
  \label{tbl:single}
  \begin{center}
    \begin{tabular}{llccc}
      \hline
      & CNN & & \multicolumn{2}{c}{FRR}\\
      \cline{4-5}
      \multirow{-2}{*}{Trait} & + Loss & \multirow{-2}{*}{EER} & @FAR0.1 & @FAR0.01\\
      \hline
      Face & \underline{R+A} & 5.636 & {\bf 20.479} & {\bf 33.033}\\
           & S+C & 5.386 & 26.792 & 40.484\\
           & S+A & {\bf 5.290} & 44.542 & 53.180\\
      \hline
      Periocular & R+A & {\bf 4.747} & 26.507 & 35.074\\
                 & \underline{S+C} & 4.836 & {\bf 19.839} & {\bf 33.721}\\
                 & S+A & 6.415 & 31.433 & 41.291\\
      \hline
      Iris & R+A  & 11.618 & 53.654 & 82.223\\
           & \underline{S+C}  & {\bf 9.095} & {\bf 36.047} & {\bf 49.241}\\
           & S+A  & 18.916 & 83.626 & 96.156\\
      \hline
      Nose & R+A & 2.804 & \textbf{9.611} & \textbf{16.255}\\
           & \underline{S+C} & {\bf 2.719} & 12.838 & 24.608\\
           & S+A & 4.124 & 13.242 & 18.201\\
      \hline
      Eyebrow & R+A & 3.316 & {\bf 15.116} & {\bf 26.602}\\
              & \underline{S+C} & \textbf{2.642} & 16.137 & 27.171\\
              & S+A & 3.654 & 20.812 & 34.362\\
      \hline
    \end{tabular}
  \end{center}
\end{table}
\begin{table*}[!t]
  \caption{Results of multibiometrics using score-level fusion, where the bold numbers indicate the best for each combination.}
  \label{tbl:score}
  \begin{center}
    \begin{tabular}{ccccc|ccc}
      \hline
      \multicolumn{5}{c|}{Trait} & \multirow{2}{*}{EER [\%]} & \multicolumn{2}{c}{FRR [\%]}\\
      \cline{1-5} \cline{7-8}
      Face & Periocular & Iris & Nose & Eyebrow & & @FAR0.1\% & @FAR0.01\%\\
      \hline
      0.5 & 0.5 & --- & --- & --- & 1.760 & 11.319 & 17.062\\
      0.5 & --- & 0.5 & --- & --- & 3.027 & 14.713 & 25.344\\ 
      0.4 & --- & --- & 0.6 & --- & 1.295 & 6.004 & 11.153\\
      0.4 & --- & --- & --- & 0.6 & 2.231 & 7.143 & 15.567\\
      --- & 0.6 & 0.4 & --- & --- & 3.528 & 13.645 & 18.296\\
      --- & 0.4 & --- & 0.6 & --- & {\bf 0.793} & {\bf 3.109} & 7.546\\
      --- & 0.5 & --- & --- & 0.5 & 1.757 & 5.363 & 12.174\\
      --- & --- & 0.4 & 0.6 & --- & 1.563 & 6.407 & 10.631\\
      --- & --- & 0.5 & --- & 0.5 & 1.699 & 7.000 & 16.327\\
      --- & --- & --- & 0.6 & 0.4 & 1.019 & 3.227 & {\bf 5.482}\\
      \hline
      0.4 & 0.3 & 0.3 & --- & --- & 1.431 & 9.729 & 15.496\\
      0.1 & 0.3 & --- & 0.6 & --- & 0.696 & 2.871 & 6.692\\
      0.3 & 0.5 & --- & --- & 0.2 & 1.384 & 6.478 & 12.340\\
      0.3 & --- & 0.3 & 0.4 & --- & 0.979 & 3.251 & 6.407\\
      0.2 & --- & 0.5 & --- & 0.3 & 1.522 & 7.262 & 12.648\\
      0.1 & --- & --- & 0.6 & 0.3 & 0.971 & 3.322 & 5.339\\
      --- & 0.3 & 0.2 & 0.5 & --- & 0.719 & 2.254 & 5.411\\
      --- & 0.2 & 0.4 & --- & 0.4 & 1.309 & 3.916 & 9.397\\
      --- & 0.3 & --- & 0.4 & 0.3 & {\bf 0.463} & {\bf 1.590} & {\bf 4.105}\\
      --- & --- & 0.4 & 0.3 & 0.3 & 0.485 & 2.515 & 5.695\\
      \hline
      0.2 & 0.2 & 0.1 & 0.5 & --- & 0.693 & 1.685 & 6.122\\
      0.2 & 0.3 & 0.2 & --- & 0.3 & 1.105 & 3.844 & 7.546\\
      0.0 & 0.3 & --- & 0.4 & 0.3 & 0.463 & 1.590 & 4.105\\
      0.0 & --- & 0.4 & 0.3 & 0.3 & 0.485 & 2.515 & 5.695\\
      --- & 0.1 & 0.3 & 0.3 & 0.3 & \textbf{0.337} & \textbf{1.495} & \textbf{3.251}\\
      \hline
      0.0 & 0.1 & 0.3 & 0.3 & 0.3 & \textbf{0.337} & \textbf{1.495} & \textbf{3.251}\\
      \hline
    \end{tabular}
  \end{center}
\end{table*}
\begin{table*}[!t]
  \caption{Results of multibiometrics using feature-level fusion, where the bold numbers indicate the best for each combination.}
  \label{tbl:feature}
  \begin{center}
    \begin{tabular}{cccccccc}
      \hline
      \multicolumn{5}{c}{Trait} & \multirow{2}{*}{EER [\%]} & \multicolumn{2}{c}{FRR [\%]}\\
      \cline{1-5} \cline{7-8}
      Face & Periocular & Iris & Nose & Eyebrow & & @FAR0.1\% & @FAR0.01\%\\
      \hline
      $\surd$ & $\surd$ & --- & --- & --- & 7.489 & 30.778 & 42.572\\
      $\surd$ & --- & $\surd$ & --- & --- & 15.916 & 71.239 & 83.056\\ 
      $\surd$ & --- & --- & $\surd$ & --- & 2.633 & 12.814 & 19.696\\
      $\surd$ & --- & --- & --- & $\surd$ & 3.590 & 13.075 & 21.974\\
      --- & $\surd$ & $\surd$ & --- & --- & 12.586 & 54.438 & 70.069\\
      --- & $\surd$ & --- & $\surd$ & --- & 1.992 & 9.967 & 18.889\\
      --- & $\surd$ & --- & --- & $\surd$ & 2.654 & 13.787 & 26.341\\
      --- & --- & $\surd$ & $\surd$ & --- & 4.911 & 31.728 & 45.444\\
      --- & --- & $\surd$ & --- & $\surd$ & 6.414 & 35.928 & 52.421\\
      --- & --- & --- & $\surd$ & $\surd$ & {\bf 1.851} & {\bf 4.865} & {\bf 7.333}\\
      \hline
      $\surd$ & $\surd$ & $\surd$ & --- & --- & 12.142 & 58.851 & 72.188\\
      $\surd$ & $\surd$ & --- & $\surd$ & --- & {\bf 1.781} & 9.516 & 14.642\\
      $\surd$ & $\surd$ & --- & --- & $\surd$ & 2.376 & 12.506 & 22.164\\
      $\surd$ & --- & $\surd$ & $\surd$ & --- & 4.744 & 26.863 & 39.203\\
      $\surd$ & --- & $\surd$ & --- & $\surd$ & 4.690 & 30.897 & 55.007\\
      $\surd$ & --- & --- & $\surd$ & $\surd$ & 2.042 & 5.529 & {\bf 7.665}\\
      --- & $\surd$ & $\surd$ & $\surd$ & --- & 5.791 & 31.870 & 43.901\\
      --- & $\surd$ & $\surd$ & --- & $\surd$ & 5.227 & 33.318 & 59.326\\
      --- & $\surd$ & --- & $\surd$ & $\surd$ & 1.637 & {\bf 8.685} & 16.255\\
      --- & --- & $\surd$ & $\surd$ & $\surd$ & 3.735 & 27.290 & 44.186\\
      \hline
      $\surd$ & $\surd$ & $\surd$ & $\surd$ & --- & 6.328 & 34.006 & 49.051\\
      $\surd$ & $\surd$ & $\surd$ & --- & $\surd$ & 5.868 & 32.677 & 47.698\\
      $\surd$ & $\surd$ & --- & $\surd$ & $\surd$ & {\bf 1.397} & {\bf 9.112} & {\bf 17.205}\\
      $\surd$ & --- & $\surd$ & $\surd$ & $\surd$ & 2.118 & 16.279 & 38.443\\
      --- & $\surd$ & $\surd$ & $\surd$ & $\surd$ & 2.789 & 20.242 & 35.999\\
      \hline
      $\surd$ & $\surd$ & $\surd$ & $\surd$ & $\surd$ & {\bf 2.797} & {\bf 19.767} & {\bf 30.636}\\
      \hline
    \end{tabular}
  \end{center}
\end{table*}
\begin{table}[!t]
  \centering
  \caption{Comparison with conventional methods and the proposed method.
  The units for EER and FRR are percentages.
  (F: Face, P: Periocular, I: Iris, LI: Left Iris, RI: Right Iris, N: Nose, E: Eyebrow)}
  \label{tbl:comp}
  \begin{tabular}{ccccc}
    \hline
    \multirow{2}{*}{Method} & \multirow{2}{*}{Trait} & \multirow{2}{*}{EER} & \multicolumn{2}{c}{FRR}\\
    \cline{4-5}
    & & & @FAR0.1 & @FAR0.01\\
    \hline
    Verma \cite{Verma-ICIP-2016} & P+I & --- & 38.5 & ---\\ 
    Ammour \cite{Ammour-TSP-2018} & F+LI+RI & 0.24 & --- & ---\\ 
    Wang \cite{Wang-TIFS-2020} & P+I & 2.29 & --- & 13.7\\ 
    Tonosaki \cite{Tonosaki-APSIPA-2023} & P+I & 1.13 & --- & 11.2\\ 
    \hline
    Proposed & P+I+N+E & 0.34 & 1.50 & 3.25\\
    \hline
  \end{tabular}
\end{table}

\subsection{Single Biometrics}

First, we evaluate the recognition accuracy of the proposed method on a single biometric trait.
To confirm the effectiveness of the combination of CNN and loss function employed in the proposed method, we use ResNet-18 and SE-ResNet-18 as CNNs, and ArcFace and CE as loss functions.
Since ResNet-18 has been demonstrated to be effective in face recognition \cite{arcface}, we use only ArcFace as the loss function.
Table \ref{tbl:single} shows the experimental results for a single biometric trait.
When EER is selected as the first priority, SE-ResNet-18+CE has the highest recognition accuracy for iris, nose, and eyebrow recognition.
For face recognition, SE-ResNet-18+ArcFace has the lowest EER, however, we select ResNet-18+ArcFace, which has almost the same EER and lower FRRs.
For periocular recognition, we use SE-ResNet-18+CE instead of ResNet-18+ArcFace, as is the case with face.
As mentioned above, EERs for iris recognition are higher than those of the other methods because of the blurred images.
Face recognition has high EERs since the whole face is not captured in the image.
On the other hand, nose and eyebrow recognition are more accurate.
This is because they are extracted more stably than other biometric traits, and they have sufficient individuality.

\subsection{Multibiometrics}

Next, we evaluate the recognition accuracy of the proposed method on two to five biometric traits.
To confirm the effectiveness of the score-level fusion, we compare its performance with feature-level fusion.
In feature-level fusion, we concatenate the features extracted from each biometric trait and produce a 1,024 dimensional feature through two fully-connected layers.
Table \ref{tbl:score} shows the experimental results for the proposed method with score-level fusion using two to five biometric traits.
The best combination for two biometric traits is periocular and nose, which improves EER up to 0.793\%.
In particular, the lowest FRR@FAR0.1\% for a single biometric trait is 9.611\%, while the combination of two improves it to 3.109\%.
The best combination of three is periocular, nose, and eyebrow, which improves EER up to 0.463\%, FRR@FAR0.1\% up to 1.590\% and FRR@FAR0.01\% up to 4.105\%.
The best combination of the four is to exclude face, which improves EER up to 0.337\%, FRR@FAR0.1\% up to 1.495\%, and FRR@FAR0.01\% up to 3.251\%.
The combination of all five gives the same result as the combination of the four, since the weight for face is 0.
The interesting point is that the use of facial parts with individual characteristics, such as the periocular, nose, and eyebrow, is sufficient to recognize a person without the use of the entire face.
Table \ref{tbl:feature} shows the experimental results for the proposed method with feature-level fusion using two to five biometric traits.
By increasing the number of biometric traits combined, the recognition accuracy does not necessarily improve, and even in the case of the best improvement, EER is 1.397\%, which is higher than that of score-level fusion.
There is a possibility that the accuracy can be improved to the same level as that of score-level fusion by modifying the approach of feature-level fusion used in this experiment, however, we consider that it is not cost-effective compared to score-level fusion.

\subsection{Comparison with Conventional Methods}

There is no method that combines the five biometric traits from a face image as the proposed method.
Therefore, we compare the proposed method with the methods used to evaluate multibiometrics in the CASIA Iris Distance database.
Note that the experimental conditions are different from those of the conventional methods.
In particular, for iris and periocular recognition, the right eye is flipped to the left eye, i.e., the number of data is twice.
Table \ref{tbl:comp} shows the comparison results between the conventional methods and the proposed method.
Verma \cite{Verma-ICIP-2016} has low recognition accuracy due to its combination of hand-crafted features.
The other methods use CNN and achieve high recognition accuracy.
Considering that the number of iris and periocular data is twice for the conventional methods, the proposed method is demonstrated to have high recognition accuracy.

\section{Conclusion}

In this paper, we proposed a novel multibiometrics method that combines face, iris, periocular, nose, and eyebrow, that can be extracted from a single face image.
This method does not sacrifice the convenience of biometrics since only a single face image is used as input.
Through a set of experiments using the CASIA Iris Distance database, we demonstrated that the method with score-level fusion of the four biometric traits except face has the highest recognition accuracy and also demonstrated the effectiveness of the proposed method by comparing it with conventional methods.

\bibliographystyle{IEEEbib}
\bibliography{paperlist}











\end{document}